\documentclass[letterpaper, 10 pt, conference]{ieeeconf}  

\IEEEoverridecommandlockouts                              

\usepackage{times} 
\usepackage{amsmath} 
\usepackage{amssymb}  
\usepackage{algorithm}
\usepackage[noend]{algpseudocode}

\usepackage{subfig}

\usepackage{booktabs}
\usepackage{pgfplotstable}

\title{\LARGE \bf
Reasoning Over Robust Controller Chains \\For Robotic Manipulation Tasks
}

\title{DC$^3$: Dynamic Control Chain Coordination for Sequential Manipulation}

\title{\LARGE \bf
FC$^3$: Feasibility-Based Control Chain Coordination}

\author{Jason Harris$^1$ \and Danny Driess$^{1,2}$ \and Marc Toussaint$^{1,2}$%
\thanks{$^1$Learning \& Intelligent Systems Lab, TU Berlin}%
\thanks{$^2$Science of Intelligence Excellence Cluster, TU Berlin}%
\thanks{~{\tt\small harris.e.jason@gmail.com}}%
\thanks{The research has been supported by the German Research Foundation (DFG) under Germany’s Excellence Strategy EXC 2002/1–390523135 ``Science of Intelligence''}%
}

\begin{document}
\maketitle
\thispagestyle{empty}
\pagestyle{empty}

\begin{abstract}
Hierarchical coordination of controllers often uses symbolic state representations that fully abstract their underlying low-level controllers, treating them as ``black boxes'' to the symbolic action abstraction.
This paper proposes a framework to realize robust behavior, which we call Feasibility-based Control Chain Coordination (FC$^3$).
Our controllers expose the geometric features and constraints they operate on. 
Based on this, FC$^3$ can reason over the controllers' feasibility and their sequence feasibility.
For a given task, FC$^3$ first automatically constructs a library of potential controller chains using a symbolic action tree, which is then used to coordinate controllers in a chain, evaluate task feasibility, as well as switching between controller chains if necessary.
In several real-world experiments we demonstrate FC$^3$'s robustness and awareness of the task's feasibility through its own actions and gradual responses to different interferences.
\end{abstract}





\section{Introduction}
The field of Task and Motion Planning (TAMP) has developed powerful tools to solve sequential manipulation problems by combining symbolic reasoning with continuous motion planning \cite{garrett2021integrated}.
However, the result of those approaches is typically an open-loop plan.
Robustly solving a task would instead require a reactive policy that switches plans or chooses different symbolic actions depending on the current situation and reacting to unexpected outcomes.
Designing such a policy goes beyond generating a single plan, but requires a form of reactive planning or contingency planning.

Paxton et al.\ \cite{paxton_representing_2019} have recently proposed a framework called Robust Logical-Dynamical System (RLDS) to generate such reactive policies. 
Our approach is related to this, but differs in how we first create a library of plans as a basis for the reactive policy, and in particular in how discrete decision making interacts with the underlying controllers to form a hierarchical policy.

Basic hierarchical control approaches introduce action abstractions, called options, together with initiation and termination conditions, but the concrete underlying controller is typically assumed a black box. 
E.g., an option can be defined by the initiation set, termination set, and a policy.

By black box we mean that the reasoning mechanisms on the higher level do not need to know any internals of the policy other than the abstract specification in terms of the initiation and termination sets.
Similarly, action operators in the Stanford Research Institue Problem Solver (STRIPS) \cite{fikes_strips_1971} and Planning Domain Definition Language (PDDL) \cite{aeronautiques1998pddl} describe the nature of a controller only in terms of abstract preconditions and effects.
What exactly the action’s underlying controller does is not exposed -- and does not need to be exposed -- to the abstract reasoning level.
Similarly, if Finite State Machines (FSM) or Behavior Trees (BT) are used to coordinate sub-level controllers \cite{correll_analysis_2018, paxton_costar_2017, hannaford_behavior_2018,colledanchise_towards_2019}, it is sufficient to have rather abstract specifications of pre-, post- and switching conditions of the controllers they coordinate (or semantic attachments).
Of course, a formalism that treats low-level controllers as black boxes is highly desirable from the perspective of the generality: They can in principle coordinate any sub-policies in any contexts. On the other hand, opening the black box can enable more informed decision making on how to coordinate sub-policies to overcome possible perturbations.

\begin{figure}
  \centering
  \subfloat[Tower scenario]{\includegraphics[width=3.9cm]{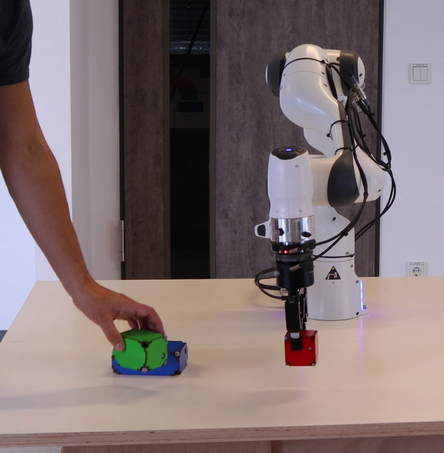}\label{fig:firstPage:tower}}
  \hspace{0.2cm}
  \subfloat[Stick scenario]{\includegraphics[width=3.9cm]{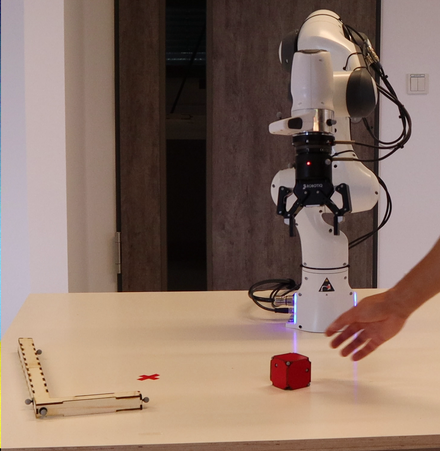}\label{fig:firstPower:stick}}\\
  \subfloat[Handover]{\includegraphics[width=8.2cm]{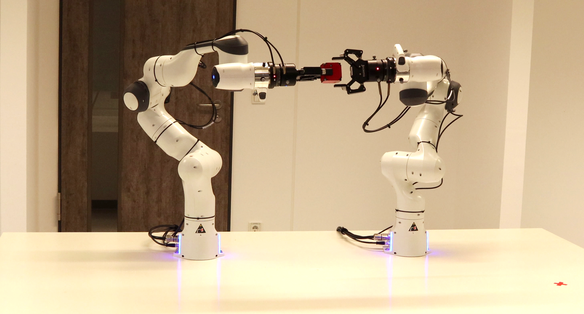}\label{fig:firstPage:handover}}
  \caption{Our system (FC$^3$) is able to overcome several perturbations in different scenarios by adapting its strategy/plan. This includes replacing a block to complete a tower, using an L-shaped stick to pull an object to a goal, or doing a hand over to bring a block to its desired position.}
  \label{fig:firstPage}
\end{figure}

We propose to use a particular kind of low-level controllers that are related to establishing geometric\footnote{We continue using the term ’geometric’ as in Logic-Geometric Programming \cite{Toussaint2015LogicGeometricPA} -- but it more generally means: defined in terms of any differentiable feature over sequences of consecutive world configurations.} constraints that are defined via differentiable features over world configurations. 
More specifically, controllers are defined as mathematical programs to achieve or maintain geometric constraints. 
The benefit of choosing these controllers is that higher level reasoning and coordination does have access to these differentiable features, and can evaluate them to estimate feasibility or bounds. 
In particular, the features can be used directly to make geometrically-informed decisions on how to coordinate the controllers, e.g. which ones can currently be executed and if they can be sequenced to achieve long-term constraints -- in contrast to merely symbolically informed higher-level reasoning or coordination systems.
The low-level controllers are fully transparent to the high-level reasoning and coordination, rather than being black boxes. We call the resulting system FC$^3$ for Feasibility-based Controller Chain Coordination.

\section{Related Work}
Finite State Machines (FSM) and Behavior Trees (BT) are typical approaches to realize reactive behavior in robotics \cite{correll_analysis_2018}. They excel in generality and are technically able to realize any strategy. Especially BTs have recently gained more recognition because of their simplicity and readability with graphical visualization \cite{paxton_costar_2017, hannaford_behavior_2018,colledanchise_towards_2019}. The generality of these systems introduces often leads to treating the underlying control policies as black-boxes. While many rely on a sufficient symbols \cite{konidaris_skills_2018} to ensure robust behavior, we propose controller chain coordination based on geometric feasibility with FC$^3$.

Paxton et. al. \cite{paxton_representing_2019} present an impressive robust system, called Robust Logical-Dynamical System (RLDS). Most behavior models (FSM and BT) require manually defined behaviors and hierarchies to achieve robust behaviors. The RLDS, however, can completely be constructed from a STRIPS-style planner \cite{fikes_strips_1971} in an automatic fashion. The robust execution of controller chains and implicit features in the present work is derived from RLDS. 
Nevertheless, the RLDS is another example of black-boxing, since the ``low-level control policies have there own convergence guarantees'' and the system does not worry about the ``motion planning part of the problem''. In several experiments we compare FC$^3$ to a system which mimics the behavior of a RLDS. We demonstrate how these assumptions and block-boxing may disrupt robust behavior in a dynamic environment. 

In Hierarchical Reinforcement Learning (HRL) an agent learns to select predefined sub-tasks, or options, instead of high-dimensional action values as in classical RL. A trained model can realize complex behavior \cite{dietterich_hierarchical_1999}, \cite{stulp_hierarchical_2011}, \cite{yang_hierarchical_2021} and does not require manual design. Options allow the agent to consider only symbolic actions during the training process. Similar to FSMs and BTs, an option is commonly fully defined by the initiation and termination conditions, and a policy. 
Options are often also treated as black boxes and therefore, low-level controller informed high-level reasoning is challenging during execution. 

Deimel \cite{deimel_reactive_2019} addresses the problem of black-boxing in reactive hierarchical behavior models by introducing a phase-state-machine. Instead of having instantaneous transitions between states, the dynamical system may timely stretch transitions, have phases, and even execute multiple transition at once. With these properties the state-phase-machine allows continuous decision processes during execution based on the information of the low-level motion generator.

Several works \cite{Toussaint2015LogicGeometricPA,wolfe2010combined,Silva2013TowardsCH,erdem_combining_2011, migimatsu_object-centric_2020} combine symbolic and motion planners to develop Task and Motion Planning (TAMP) solvers. These approaches can solve complex problems, but would require full re-planning when perturbations occur.
The experiments in \cite{Silva2013TowardsCH} and \cite{gharbi_combining_2015} take several seconds to find a solution in the geometric space.
Those solutions rely on motion trajectories, which may fail when interference and perturbations occur. The framework presented in the present work also relies on the combination of symbolic and geometric information. However, instead of solving for an entire motion trajectory, FC$^3$ inspects sequence feasibility of sequential controllers in real-time suitable manner. Therefore, we also call our approach TAMP execution. Migimatsu et al. \cite{migimatsu_object-centric_2020} also build a framework for TAMP execution but do not consider switching plans and therefore would require full replanning to overcome severe disturbances.

Ratliff et al. \cite{ratliff_riemannian_2018} presents a framework for combining multiple motion policies to a motion generator called Riemannain Motion Policy (RMP). Each policy in the RMP is a contributor to a continuously generated motion. Each contributor is paired with a Riemann metric, which accounts for its local geometry (the directions of importance to the policy). The combined motion of the RMP is achieved through the metric weighted average of these policies and considering their Riemann metric. Many works have utilized RMPs for the underlying low-level controller in hierarchical approaches \cite{paxton_representing_2019,meng_neural_2019,wingo_extending_2020}. The work in \cite{cheng_rmpflow_2019} presents how multiple RMPs can be be combined to a global policy.

Our constrained operational space controller presented in this work has some similarities to the RMP framework. Each feature and constraint of the controller contributes to the final motion policy. The resulting controller can therefore be also considered an ensemble of controllers or a motion generator. 

\section{Controller Chains \& Feasibility}

\subsection{Action Controller}

For each (macro) action we have a specific controller $\Omega$. A controller defines both, a control law (determining the desired reference acceleration of the robot) as well as feasibility criteria which enable high-level coordination and reasoning with these. Controllers thereby provide the fundamental building blocks for FC$^3$. In this section we detail our conventions to define controllers and their feasibility.

\subsubsection{1-Step Constrained Operational Space Controller}

Let $\mathcal{X}$ be the configuration space of the scene, including both robot and object degrees of freedom. We assume a controller that defines the next reference configuration $x' = x(t+\tau)\in\mathcal{X}$ at the given state $\hat x = (x(t),\dot x(t))$ in terms of a non-linear mathematical program
\begin{align}
    \min_{x'}\quad&c(\hat x, x') + \phi(\hat x,x')^T\phi(\hat x, x') \nonumber\\
    \text{s.t.}\quad&g(\hat x,x') \le 0. \label{eq:ctrl}
\end{align}
For small $\tau$ this becomes equivalent to optimizing over the reference acceleration; however, to define transient constraints below it is more convenient to formulate the problem as a 1-step optimization over the next configuration $x' = x(t+\tau)$.
Neglecting the hard (in)equality constraints, this can be viewed as an instance of general operational space control \cite{peters_unifying_2008}, where $c$ defines control costs, and features $\phi(\hat x,x') \in\mathbb{R}^d$ define a square cost in a $d$-dimensional feature space (e.g.\ end-effector space). In our case, control costs will penalize square accelerations $c(\hat x, x') = \alpha^2/\tau^2 |\!|x'-(x+\tau\dot x) |\!|^2$, which can also be subsumed in the feature vector $\phi$ in addition to other control objectives such as endeffector position, orientation, alignments, etc.

We straight-forwardly extend the formulation to include hard control constraints, represented by inequalities $g$. For brevity of notation this includes equality constraints (which could be viewed as double inequality constraints but are handled differently by the underlying solver). In practice, we will often specify endeffector position and alignment objectives as hard constraints.

We leverage an efficient Augmented Lagrangian constrained nonlinear program (NLP) solver with a Newton method in the inner loop to solve (\ref{eq:ctrl}) in each control cycle. Note that the first Newton step direction of this solver coincides with the standard operational space control solution using a local linearization of the control objectives $\phi$. Further iterations and the outer Augmented Lagrangian loop ensure (local) convergence to exact constraint fulfillment for nonlinear $\phi$ and $g$.

\subsubsection{Immediate and Transient Features}

During the execution of an action, some constraints should hold immediately and throughout, while other constraints should hold at the end of the action execution, e.g.\ when reaching for a desired position. We call the latter kinds of constraints \emph{transient}. More precisely, assume a transient feature/constraint depending only on $x'$ currently has a significantly non-zero ``error'' $\phi(x_0) = a$, but at termination of this action should reach the target $\phi(x_T) = 0$. To realize this using the above controller formulation we define a shifting feature $\hat\phi$ which relates to a constant velocity in feature space, namely by clipping the step length in error reduction:
\begin{align}
\hat\phi(x) = \max\{1, \tau\epsilon/|a|\}~ \phi(x) ~.
\end{align}
Here, $\epsilon$ has the semantics of a maximal velocity in feature space to move from $a$ to zero.

In summary, the controlled motion is fully specified by a tuple $(\phi, g, \epsilon)$, where $\epsilon_i \in \mathbb{R}$ indicates for each entry of the feature $\phi$ and $g$ whether it is immediate ($\epsilon_i=\infty$) or transient ($\epsilon_i$ defines the clipping feature velocity).

We say a controller has converged if both, immediate and final transient constraints are met (with a threshold in practice). 

\subsubsection{Discrete Action Signal}

After convergence of a controller, a discrete control signal $\psi$ such as opening or closing the gripper may be appended. We assume $\psi \in\{\psi_G, \psi_P, \texttt{nil}\}$ for grasping ($\psi_G$), placing ($\psi_P$) and no action (\texttt{nil}). The discrete signal $\psi$ has no influence on the 1-step controller NLP (\ref{eq:ctrl}), but plays a role for solving for sequence feasibility in a chain of controllers.

To finalize, a controller is fully specified by a tuple
\begin{align}
    \Omega = (\phi, g, \epsilon, \psi) ~.
\end{align}

\subsection{Feasibility of a Controller}
Given a candidate controller $\Omega = (\phi, g, \epsilon, \psi)$ we can predict its feasibility, costs, and execution time.
Specifically, for a given configuration $x \in \mathcal{X}$, we can determine the immediate feasibility $F_I(\Omega, x)$, where we only take the hard constraints $g$ into account. 
We say a controller $\Omega$ is feasible in a configuration $x$, when $F_I(\Omega, x)$ holds.

We can also determine the final feasibility $F_T(\Omega, x)$ by considering all features, but removing any control cost penalties.
A controller $\Omega$ has converged when $F_T(\Omega, x)$ holds for a given configuration $x$. In relation to options or operators, $F_I(\Omega, x)$ and $F_T(\Omega, x)$ are indicators for the initiation and termination sets, respectively. $F_I(\Omega, x)$ and $F_T(\Omega, x)$ are indicators for initiation and termination during execution by considering the current configuration $x$. For convenience, we use $F_I(\Omega)$ and $F_T(\Omega)$ for a current given configuration $x$. 

Furthermore, we can solve for a future point of convergence by considering NLP $(\mathcal{X}, \phi_\Omega, g_\Omega)$. The decision variable $x_T \in \mathcal{X}$ describes a configuration in a far future, in which $F_T(\Omega, x_T)$ holds.  

The same NLP also allows us to estimate control costs. In fact, by solving for a whole path (with potentially coarse time discretization) from the current state to a point of final convergence we can compute lower bounds on the control costs (and execution time) with variable fidelity. Note that all our predictions over immediate feasibility, final feasibility, and the control costs can be made strict lower bounds\footnote{Also in the sense of infeasibility} of the true control costs neglecting stochasticity. As always in decision making, such lower bounds are valuable pieces of information of higher-level coordination of such sub-routines.

\subsection{Controller Chains}
A chain of controllers $\mathbf{\Omega}=(\Omega_1,...,\Omega_N)$ defines the sequence of motions and manipulation actions FC$^3$ attempts to complete a task. In addition to the control chain we have a set of goal constraint functions $G$, i.e., we require $G(x)\le 0$ for goal configurations $x\in\mathcal{X}$. By evaluating $F_I(G)$, FC$^3$ can indicate if a task has been fulfilled, i.e.\ the goal has been reached.

\subsubsection{Augmentation with Implicit Constraint}

The robust execution of controller chains is based on the selection of the most downstream and feasible controller in the chain. 

Whenever a controller is selected and converges, the subsequent controller is expected to be feasible.
At convergence of the last controller we want the task constraints $G$ to be fulfilled. 
In order to achieve this, we follow the approach of RLDS \cite{paxton_representing_2019} to compute additional \emph{implicit} constraints by back-propagation through the chain, as described in the following.

Let $\Omega_i$ be a controller in a controller chain $\mathbf{\Omega}$.
By evaluating final feasibility $F_T(\Omega_i)$ of the controller $\Omega_i$ we can also solve for a configuration $x_i$ that fulfills these terminal constraints of $\Omega_i$.
The implicit constraints are derived by identifying the immediate constraints in $g_{\Omega_{i+1}}$ of the subsequent controller $\Omega_{i+1}$ which do not hold in $x_i$. Specifically, we augment the constraints of controller $\Omega_i$ with those constraints functions of $\Omega_{i+1}$ that do not hold at $x_i$:
\begin{align}
    I &= \{d : [g_{\Omega_{i+1}}(x_i)]_d > \epsilon \} \nonumber\\
    g_{\Omega_i} &\gets g_{\Omega_i} \cup [g_{\Omega_{i+1}}]_I ~, \label{eq:implicit_features} 
\end{align}
where $I$ are those constraint functions within $g_{\Omega_{i+1}}$ that are violated (by a margin $\epsilon$) at $x_i$. Implicit constraints represent the subgoal/milestones, which are not originally achieved by controller $\Omega_i$, but are necessary to initiate $\Omega_{i+1}$. 
This backward recursion of propagating implicit constraints is started by defining $g_{\Omega_{N+1}} \equiv G$ to be the task constraints. The implicit features assure that FC$^3$ enters the most efficient\footnote{closest to the goal} downstream controller during the execution of a controller chain.

\subsubsection{Sequence Feasibility}

Consider two controllers $\Omega_1$ and $\Omega_2$. We want to analyze the feasibility of chaining these. Their sequence feasibility is given, when $F_T(\Omega_1)$ and $F_I(\Omega_2)$ hold in the same configuration $x_s$, which we call a switching configuration.
We can solve for a potential switching configuration with the NLP that includes all constraints of $\Omega_1$ (as for $F_T(\Omega_1)$, but only the immediate features of $\Omega_2$ (as for $F_I(\Omega_2)$). Solving a switching configuration implies sequence feasibility.

This can be generalized to an entire chain $\Omega_{1:N+1}=(\Omega_1,...,\Omega_N, G)$. 
We can determine the feasibility of a chain, including the goal controller, to reach a goal by solving the NLP
\begin{align}
     \min_{x_{0:N+1}}\quad&\sum\limits_{i=0}^{N+1} \phi_i(x_i)^T \phi_i(x_i) \nonumber\\
    \text{s.t.}\quad
    &\forall_{i\in N}: g_{\Omega_i}(x_i) \le 0 \nonumber \\
    &\forall_{i\in N}: g_{\Omega_{i+1}}(x_i) \le 0
\label{eq:KOMO_plan_feasibility}
\end{align}
where $\phi_i$ and $g_{\Omega_i}$ are defined by $\Omega_i$ in the controller chain $\Omega$.
As in LGP \cite{Toussaint2015LogicGeometricPA}, we solve this problem by augmenting all robot manipulators with a virtual manipulation frame.
A grasp action imposes a 6D positional equality and zero-velocity constraint between the virtual frame and the grasped object as part of the respective controller constraints $g_{\Omega_i}$.
When an object is placed we impose equality constraints on the object and the position where it is released.
The sequence feasibility of a controller chain is similar to the sequence bound in LGP \cite{Toussaint2015LogicGeometricPA}. We solve \eqref{eq:KOMO_plan_feasibility} with the KOMO framework \cite{toussaint_newton_2014}.

\subsubsection{Augmentation with Logic Preconditions}

The switching configurations also inform about the phases a gripper is expected to hold to an object.
As the gripper state is also a precondition for these controllers, we introduce logical preconditions $\psi' \in \{\psi_G', \psi_P'\}$ for the controllers in phases in which a gripper holds ($\psi_G'$) an object and when the gripper is expected to be free ($\psi_P'$). These implicit controller preconditions slightly modify the function $F_I(\Omega)$ to also check for logic feasibility. This ensures that a discrete action signal is executed before entering a more downstream controller in the chain.

\begin{figure}
  \centering
  \includegraphics[width=0.35\textwidth]{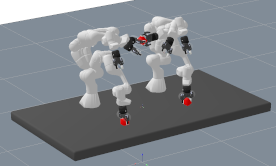}
  \caption{Visualization of the switch configurations in a 3-controller chain for a grasp, hand over, and place sequence. The switch configurations indicate during which execution phase a gripper is expected to hold the block. Both grippers are free when the left robot moves to grasp the block. While the left brings to block to the right robot, it is expected to hold the object. After the hand over, the right robot holds the block until placing it. }
  \vspace{-0.5cm}
  \label{switch_configuration_handover}
\end{figure}

\section{Feasibility-Based Controller Chain Coordination}

\subsection{Creation of a Library of Controller Chains from the Backward Action Tree}

We present how FC$^3$ generates multiple controller chains from a STRIPS-style problem formulation and a backwards tree search. 

\subsubsection{Goal Regression STRIPS planning}

In typical STRIPS planning, the planner begins at an initial state $I$ and searches the symbolic state space until the goal state $L_G$.  
A goal regression planner \cite{waldinger1981achieving, kaelbling_hierarchical_2011}, however, begins at $L_G$ and extends logical states with relevant actions backwards.
A symbolic action $a$ expands to a new logic state $L_S'$ with the function
\begin{equation}
    \label{eq:relevant_action_extension}
    \text{rel}(a,L_S) = 
    \begin{cases}
      L_S' &\text{if}\quad 
      L_S \cap \gamma_a  \neq \emptyset  \wedge 
      L_S \cap \delta_a  = \emptyset \\
      L_S & \text{otherwise}
    \end{cases},     
\end{equation}
where $\beta_a$ are the preconditions, $\gamma$ are the add-effects and $\delta$ are the delete-effects of the action $a$. The new state is derived with $L_S'=(L_S\setminus\gamma_a)\cup\beta_a$ if the action $a$ is relevant in $L_S$.

A goal regression planner may encounter unreachable states while searching the state space. 
An unreachable state $L_S$ may fulfill the termination criterion $L_I\subseteq L_S$, but the resulting forward plan $P$ will not necessarily lead to the goal state $L_G$.
For that reason it is necessary to certify/verify that a potential initial state $L_S$ also leads to the goal state $L_G$ through following the forward action chain with \Call{Forward}{$L_S$}.

\subsection{Exploring and Trimming the Action Tree}

We allow FC$^3$ to explore the search tree beyond typical termination criterion and incorporate additional actions to the action tree.
The parameter $j$ defines the maximum depth of the action tree beyond the length of the shortest verified plan starting at the initial state. The modified backward tree search is shown in Algorithm \ref{alg:FC3_backwards_tree_search}. Furthermore, we define a parameter $r$, which trims the action tree $T$ around the shortest verified plan $P$ in \Call{TrimActionTree}{$T$, $P$, $r$}. The additional parameters $j$ and $r$ allow us to shape the resulting action tree and characterize how far FC$^3$ may diverge from a primary chain.

\begin{algorithm}
\caption{Exploring and Trimming the Action Tree}
\label{alg:FC3_backwards_tree_search}
\begin{algorithmic}[1]
\Function{GenerateActionTree}{$L_G$,$L_I$,$\mathcal{A}$,$j$}
    \State \textbf{Given}: logic goal state $L_G$, logic initial state $L_I$, symbolic actions $\mathcal{A}$, $j$ explore range
    \State $q$ = \Call{PriorityQueue}{$ $}
    \State $T$ = \Call{Tree}{$ $}
    \State $P$ = $\emptyset$ //shortest path to from $L_I$ to $L_G$
    \State $d_I$ = 0 // depth of verified candidate for initial state
    \State \Call {Push}{$q$,$L_G$}
    \While{$q \neq \emptyset$}
        \State $L_S$ = \Call{Pop}{$q$}  
        \If{$P \neq \emptyset \wedge L_I\subseteq L_S \wedge \Call{Forward}{L_S} \subseteq L_G$}
            \State $P$ $\gets$ \Call{ForwardPlan}{$L_S$}
            \State $d_I$ = \Call{D}{$L_S'$}
        \EndIf
        \For{$a \in \mathcal{A}$} 
            \If{$L_S \cap \gamma_a  \neq \emptyset  \wedge L_S \cap \delta_a  = \emptyset$}
                \State $L_S' = (L_S\setminus\gamma_a)\cup\beta_a$
                \If{$L_S' \notin T$}
                    \State \Call {UpdateTree}{$T$,$S'$, child=$S$}
                    \If{$P$ = $\emptyset$ $\vee$ \Call{D}{$L_S'$} $>$ $d_I$ + $j$}
                        \State \Call {Push}{$q$,$L_S'$} // new relevant state
                    \EndIf
                \EndIf
            \EndIf
        \EndFor
    \EndWhile
    \State \Return $T, P$
\EndFunction
\end{algorithmic}
\end{algorithm}

\subsection{Reasoning over Feasibility}

Finally we present how all components are brought together to realize robust behavior for manipulation tasks. The initialization phase consists of the following steps:
\begin{enumerate}
  \item Map every symbolic action $a \in \mathcal{A}$  to a controller $\Omega$. (In generality and for convenience, one can also map a single symbolic action to a subchain of controllers.)
  \item Generate Action Tree $T$ with backwards planner as shown in Alg. \ref{alg:FC3_backwards_tree_search}.
  \item Trim the initial Action Tree $T$ in \Call{TrimActionTree}{$T$}. We remove all logical states from $T$, which have a shortest distance to any logical state in $P$ greater than $r$.
  \item Generate a list/set of controller chains in \Call{BuildControllerChains}. This is done by creating every subchain in $T'$ and replacing each symbolic action with its mapped controller chain. For $K$ nodes in $T'$ we therefore receive $K-1$ controller chains.
  \item Finally we update each controller in its chain with implicit features as shown in \eqref{eq:implicit_features}.
\end{enumerate}
It is important to note that, during the initialization, no controller chain is selected for execution. That is because FC$^3$ selects a feasible plan be reasoning over the controller chains' feasibility imposed by the action tree. During execution, FC$^3$ continuously does the following steps:
\begin{itemize}
  \item If the selected controller chain $\mathbf{\Omega}$ is empty, select a new feasible controller chain.
  \item Otherwise execute the most downstream controller $\Omega$ as long as the remaining chain $\Omega_{i:N}$ is feasible, where $i$ is the index of the last executed controller. For computational efficiency, the sequence feasibility of $\Omega_{i:N}$ is not checked in every iteration.
  \item FC$^3$ terminates, when either the goal has been reached $F_I(G)$ or no feasible plan has been found.
\end{itemize}

\begin{figure}
  \centering
  \vspace{-1.3cm}
  \includegraphics[scale=0.7]{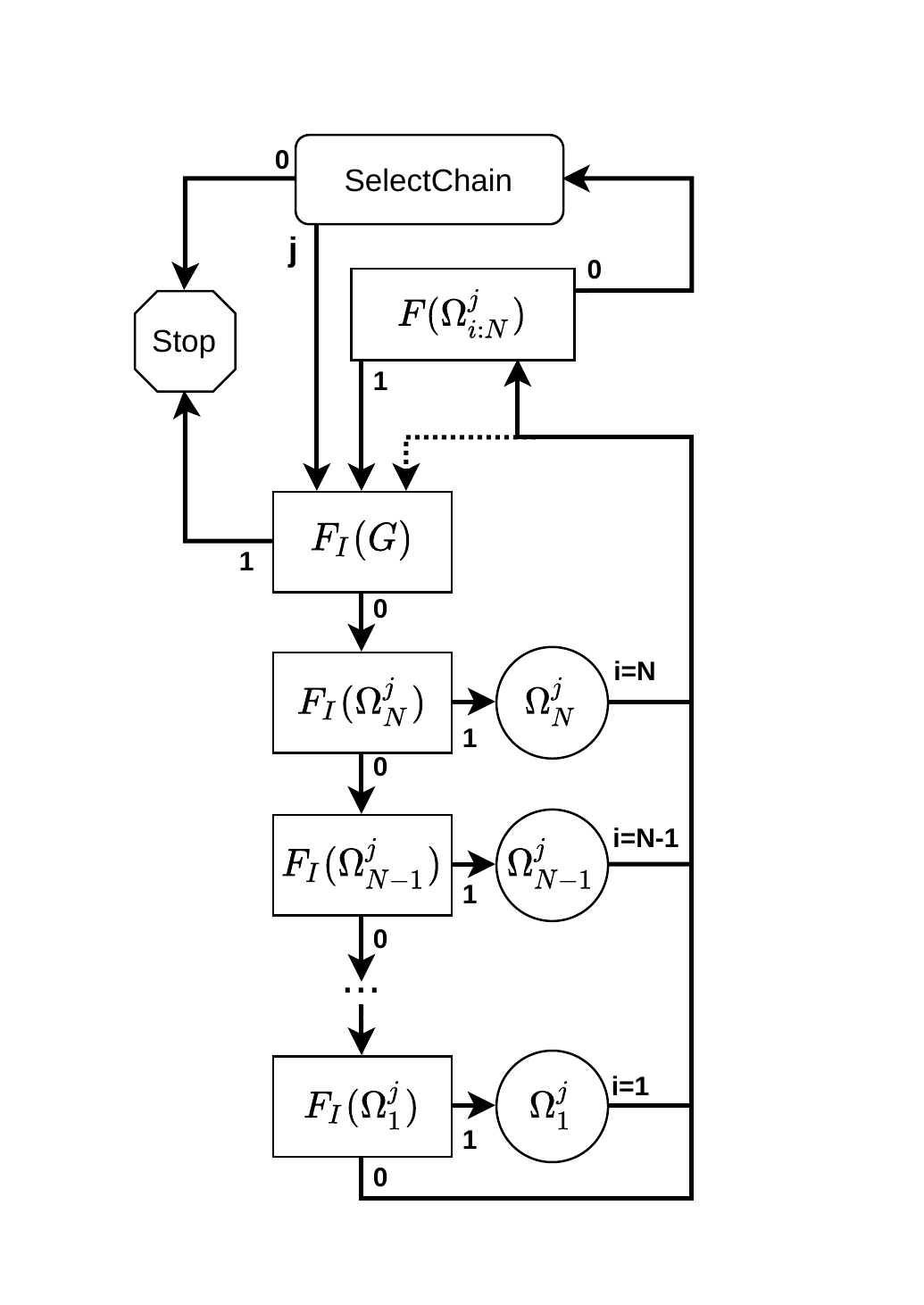}
  \vspace{-0.5cm}
  \caption{Diagram of FC3 Execution. Once a feasible controller chain has been selected, it is executed as long as the remaining chain $\Omega_{i:N}$ is feasible, where $i$ is the last entered controller. The feasibility of the remaining chain is not checked in every execution cycle, indicated with the dotted arrow. FC$^3$ stops the execution either when the goal has been reached, or no controller chain is feasible.}
  \vspace{-0.3cm}
  \label{FC$^3$_diagramm}
\end{figure}

The behavior during execution is visualized in Fig. \ref{FC$^3$_diagramm}.

\algdef{SE}[DOWHILE]{Do}{doWhile}{\algorithmicdo}[1]{\algorithmicwhile\ #1}%
\begin{algorithm}
\caption{FC$^3$ Initialisation and Execution }
\label{alg:FC3_init_and_exec}
\begin{algorithmic}[1]
\Function{FC$^3$}{$L_G$, $L_I$, $\mathcal{A}$, $G$}
    \State $(T, P)$ = \Call{GenerateActionTree}{$L_G$, $L_I$, $\mathcal{A}$}
    \State $T'$ = \Call{TrimActionTree}{$T$, $P$, $r$}
    \State $\Pi$ = \Call{BuildControllerChains}{$T'$}
    \For{$j=1,\ldots, |\Pi|$}  
        \State \Call{UpdateImplicitConstraints}{$\mathbf{\Omega}^j$, $G$}
    \EndFor
    \State $i = 0$
    \State $\mathbf{\Omega} \gets \emptyset$
    \Do
        \State // select control chain
        \If{\textbf{not} $F(\mathbf{\Omega}_{i:N})$}
            \State $\mathbf{\Omega} \gets \emptyset$
            \For{$j=1,\ldots, |\Pi|$}  
                \If{$F(\mathbf{\Omega}^j)$}
                    \State  $\mathbf{\Omega} \gets \mathbf{\Omega}^j$; \textbf{break}
                \EndIf
            \EndFor
        \EndIf
        \State // execute control chain
        \For{$i = |\mathbf{\Omega}|, \ldots, 1$}  
            \If{$F_I(\mathbf{\Omega}_i) = 1$}
                \State \Call{Step}{$\mathbf{\Omega}_i$}; \textbf{break}
            \EndIf
        \EndFor
    \doWhile{\textbf{not} $F_I(G) \vee \mathbf{\Omega} = \emptyset $}
\EndFunction
\end{algorithmic}
\end{algorithm}

\section{Experiments}
We conducted several real-robot experiments to validate the robustness of our system.
In three different scenarios one or two Franka Emika Panda 7-DoF robot arms, each equipped with a Robotiq 2F-85 gripper, attempt to complete a manipulation tasks.
During the trials, a human disturbs the robots and hinders the task. 
The objects in the environment are tracked with an Optitrack motion capture system to estimate the configuration of the scene. 
For baselines, we compare FC$^3$ to two other systems:
\subsubsection{Linear Execution}
this system executes one chain of controllers in orderly fashion. 
It executes the current controller $\Omega_i$ until $F_I(\Omega_{i+1})$ is fulfilled. It then moves to the subsequent controller $\Omega_{i+1}$ and repeats the process. The system strictly follows the controller chain and cannot jump back- or forwards in the chain.
\subsubsection{RGDS}
this system is meant to imitate the RLDS \cite{paxton_representing_2019} with our geometric controllers as underlying policy. RGDS stands for Robust Geometric-Dynamical System. The initialisation and execution in Alg. \ref{alg:FC3_init_and_exec} is modified slightly to fit the behavior of RLDS. During initialisation, a feasible plan is selected once and cannot be changed during execution. Additionally, RGDS does not observe the remaining sequence feasibility of the selected chain.

\subsection{Tower}
\label{ch:experimental_setup:tower}
In this scenario (see Fig.~\ref{fig:firstPage:tower}), the task is to build a tower by stacking three blocks with the right robot arm. 
At the start of each trial, all blocks are placed at random positions on the table and are reachable for the robot. 
One of the following interference/disturbance is done by a human in each trial:
\begin{itemize}
    \item[I0:]no interference, the robot is not disturbed in any way while stacking the blocks.
    \item[I1:]while the robot moves to grasp the green block, the block is moved, but remains reachable for the robot
    \item[I2:]while the robot closes the gripper to grasp the green block, the block is moved, but remains reachable for the robot
    \item[I3:]after the robot has placed the green block on the blue block and moves to the red block, the green block is removed from the tower and placed back on the table. The green block is still in reach.
    \item[I4:] while the robots moves to the green block, the red block is moved out of reach. The task is infeasible.
    \item[I5:]while the red block is grasped and moved to the green block, the green block is removed from the tower and placed back on the table. It is still reachable for the robot.
    \item[I6:]while the robot moves to grasp the green block, the green block is placed on the blue block.
\end{itemize}

Each interference was performed in five separate trials, resulting in 35 trials for each system. If the task was not completed after 120 seconds, the task was labeled as failed. 
The results are discussed in Sec.~\ref{ch:experiment_results} and Tab.~\ref{tab:tower_experiment_results}.

\begin{table}
  \begin{center}
  
    \begin{filecontents*}{data/tower_all.csv}
interference_num,Linear Execution,RGDS,FC$^3$
I0,32.8 $\pm$ 6.1,28.6 $\pm$ 2.6,29.5 $\pm$ 3.5
I1,101.9 $\pm$ 40.5,34.2 $\pm$ 4.8,36.4 $\pm$ 2.6
I6,83.1 $\pm$ 50.6,15.4 $\pm$ 8.5,14.8 $\pm$ 5.8
I2,-,50.2 $\pm$ 7.3,44.9 $\pm$ 4.9
I3,-,45.0 $\pm$ 5.4,42.0 $\pm$ 4.1
I5,-,-,56.1 $\pm$ 6.8
I4,-,-,11.3 $\pm$ 5.2
\end{filecontents*}
\pgfplotstabletypeset[
     font={\small},
      assign column name/.style={/pgfplots/table/column name={\textbf{#1}}},
      col sep=comma, 
    display columns/0/.style={
        column name=$ $, 
        column type={c},string type},  
     display columns/1/.style={
        column type={c},string type},  
     display columns/2/.style={
        column type={c},string type},  
    display columns/3/.style={
        column type={c},string type}, 
      every head row/.style={
        before row={\toprule}, 
        after row={\midrule},
            },
      every last row/.style={after row=\bottomrule}, 
      every row no 6/.style={
      before row={\hline}, 
      },
    ]{data/tower_all.csv} 
    \caption{Average execution time [s] in the Tower scenario. If no trial was successful, we denote it with ``-''. I4 is a perturbation which made the task infeasible for the robot. FC$^3$ did not only overcome every perturbation, but also identified infeasible tasks and terminated execution.}
    \vspace{-0.3cm}
    \label{tab:tower_experiment_results}
  \end{center}
\end{table}

\subsection{Stick Pull}
\label{ch:experimental_setup:stick_pull}
In this scenario (see Fig.~\ref{fig:firstPower:stick}), the task for the right robot arm is to move the red block to a specified goal position. 
This can either be done by grasping and placing the block, or by using an L-shaped stick to pull the block to the goal position. 
In the beginning of each trial, both objects are reachable for the robot. One of the following interference occurs in each trial:
\begin{itemize}
    \item[I0:]no interference, the robot is not disturbed while moving to the block
    \item[I1:]the red block is moved while the Panda moves to or grasps the block, but it remains in reach
    \item[I2:]the red block is moved out of reach for the robot to grasp, or to reach with the stick. The task is infeasible for the robot. 
    \item[I3:]the red block is moved out of reach while the robot moves to or grasps the block, but can be pulled to the goal position if the robot uses the stick
    \item[I4:]the red block and the stick is are moved out of reach. The task is infeasible for the robot. 
\end{itemize}
As with the former scenario, each interference was performed in five separate trials, resulting in 25 trials for each system.
If the task was not completed after 120 seconds, the task was labeled as failed. 

Tab.~\ref{tab:stick_experiment_results} shows the results that are discussed in Sec.~\ref{ch:experiment_results}.

\begin{table}
  \begin{center}
   
    \begin{filecontents*}{data/stick_all.csv}
interference_num,Linear Execution,RGDS,FC$^3$
I0,13.9 $\pm$ 0.8,15.1 $\pm$ 2.0,14.5 $\pm$ 2.4
I1,-,22.1 $\pm$ 3.7,27.6 $\pm$ 8.9
I3,-,-,32.0 $\pm$ 5.1
I2,-,-,18.0 $\pm$ 6.3
I4,-,-,22.5 $\pm$ 9.0
\end{filecontents*}
\pgfplotstabletypeset[
     font={\small},
      assign column name/.style={/pgfplots/table/column name={\textbf{#1}}},
      col sep=comma, 
    display columns/0/.style={
        column name=$ $, 
        column type={c},string type},  
     display columns/1/.style={
        column type={c},string type},  
     display columns/2/.style={
        column type={c},string type},  
    display columns/3/.style={
        column type={c},string type}, 
      every head row/.style={
        before row={\toprule}, 
        after row={\midrule},
            },
      every last row/.style={after row=\bottomrule}, 
      every row no 3/.style={
      before row={\hline}, 
      },
    ]{data/stick_all.csv} 
     \caption{Average execution time [s] in the Stick Pull scenario. If no trial was successful, we denote it with \mbox{``-''}. I2 and I4 are perturbations which made the task infeasible for the robot. FC$^3$ did not only overcome every perturbation, but also identified infeasible tasks and terminated execution.}
     \vspace{-0.5cm}
    \label{tab:stick_experiment_results}
  \end{center}
\end{table}

\begin{figure*}
  \centering
  \includegraphics[width=\textwidth]{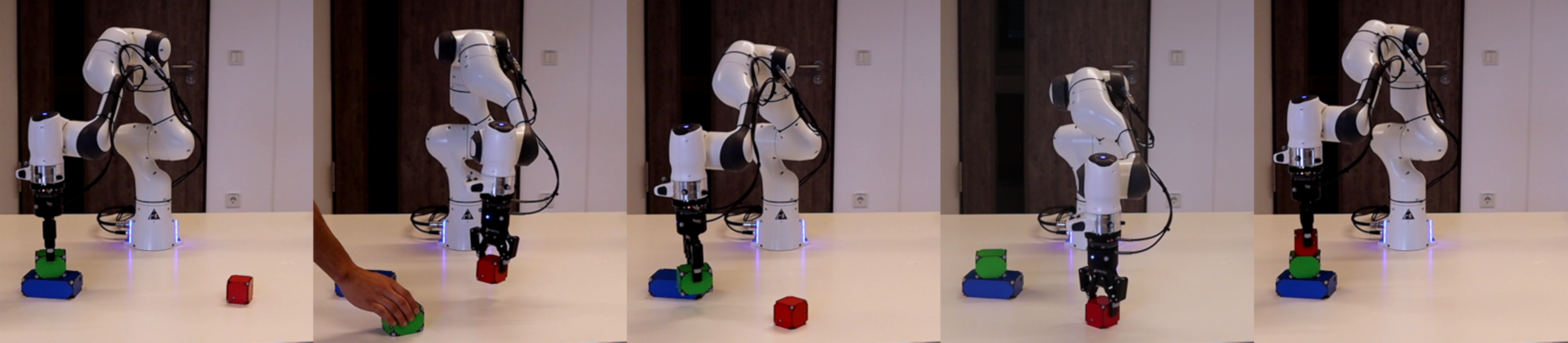}
  \caption{Snapshots of FC$^3$ Interference 5 in the tower scenario. The green block gets removed and placed back on the table, while the arm holds the red block. FC$^3$ recognizes that the current control chain is no longer feasible. It then switches to a new chain, which allows the robot to place the gripped object back on the table first to then complete the task. The linear execution and RGDS get stuck when the perturbation occurs and cannot complete the task.}
  \vspace{-0.5cm}
  \label{fig:tower_timeline}
\end{figure*}

\subsection{Handover}
\label{ch:experimental_setup:handover}
The experiments in this scenario (see Fig.~\ref{fig:firstPage:handover}) investigate the ability of FC$^3$ to determine the feasibility of the control chains, in order to switch between them during execution when disturbances occur. 
The other systems are excluded from this experiment, as they do not switch between control chains during execution. 
The task for both arms is to place the red box on the designated goal position on the table near the right robot. 
The red block is reachable by either robot.
Therefore, the task remains feasible during the experiment. 
The robots can grasp, place, or hand-over the block to the other robot arm. 
In each trial, the initial position of the block alternates with the possible interference:
\begin{itemize}
    \item[I0:] the red block's initial position is near the right arm and is not moved
    \item[I1:] the red block's initial position is near the right arm and is moved between the two robots
    \item[I2:] the red block's initial position is near the right arm and moved to a position only reachable by the left robot
    \item[I3:] the red block's initial position is near the left arm and is the block is not moved
    \item[I4:] the red block's initial position is near the left arm and is moved between the two robots
    \item[I5:] the red block's initial position is near the left arm and moved to a position only reachable by the right robot
\end{itemize}
For each case, at least one plan is feasible. For the cases I2 and I4 more than one plan is feasible, but the simplest option is for the left robot to pick-and-place the block compared for the right robot to pick, do a handover, and finally the left arm placing the block. 

\begin{table}
  \begin{center}
  
    \begin{filecontents*}{data/hand_over_all.csv}
interference_num,FC$^3$
I0,14.2 $\pm$ 2.8
I1,57.3 $\pm$ 13.6
I2,29.1 $\pm$ 6.4
I3,41.8 $\pm$ 7.4
I4,27.6 $\pm$ 7.0
I5,52.2 $\pm$ 8.8
\end{filecontents*}
\pgfplotstabletypeset[
     font={\small},
      assign column name/.style={/pgfplots/table/column name={\textbf{#1}}},
      col sep=comma, 
    display columns/0/.style={
        column name=$ $, 
        column type={c},string type},  
     display columns/1/.style={
        column type={c},string type},  
     display columns/2/.style={
        column type={c},string type},  
    display columns/3/.style={
        column type={c},string type}, 
      every head row/.style={
        before row={\toprule}, 
        after row={\midrule},
            },
      every last row/.style={after row=\bottomrule}, 
    ]{data/hand_over_all.csv} 
     \caption{Average execution time [s] in the Handover scenario. FC$^3$ was able to identify a feasible controller chain for different initial block positions and complete the task regardless of the perturbation.}
     \vspace{-0.5cm}
  \label{tab:hand_over_experiment_results}
  \end{center}
\end{table}

\subsection{Results}
\label{ch:experiment_results}

The results of the experiments for each scenario are shown in Tab.~\ref{tab:tower_experiment_results}, \ref{tab:stick_experiment_results} and \ref{tab:hand_over_experiment_results}. 
FC$^3$ and RGDS display similar robustness for interference, in which resetting the control chain, i.e. entering the most efficient controller, is sufficient. 
For those tasks that can be solved with RGDS, the execution times between our proposed FC$^3$ and RGDS as the baseline are comparable. 
However, we see that FC$^3$ is able to complete all feasible tasks, while RGDS fails to complete tasks under certain perturbations (Tower I5, Stick I3). 
This shows the importance of the ability to switch control chains. 
Not all perturbations can be overcome by resetting to a previous action, but require a change of plan.
Paxton et al.~\cite{paxton_representing_2019} present how a hierarchical RLDS could hold multiple plans, yet do not discuss the switching conditions for such a hierarchical system. Certain perturbations require the need to change the strategy, as FC$^3$ does by selecting control chains based on their geometric feasibility.

Additionally, FC$^3$ presents awareness and appropriate reaction to such perturbations that make the task infeasible (Tower I4, Stick I2, and I4).
In these situations, our system recognizes that no other control chain is feasible, and terminates the execution.
The linear execution model and RGDS fail to do so and are stuck in the execution loop. This often leads to undesired behavior of the robot, e.g.\ full arm extension or running close to collision.

In the handover scenario, FC$^3$ demonstrates the ability to identify a feasible controller for a different environment and adapt to interference. 
In every situation FC$^3$ is able to complete the task, either by simple pick-and-place or performing a handover to overcome the perturbation.
However, the interference I4 reveals that FC$^3$ not necessarily will execute the most efficient control chain. 
The perturbation moves the block between the two robots and consequently becomes reachable for both. Yet the system continues to execute the selected chain, because it remains feasible. 

While FC$^3$ demonstrates robustness over various perturbations, it also demonstrates the ability to cooperate. The interference 6 in the Tower scenario is unique from the other perturbations, because it aids the task, instead of preventing and disturbing the robot. This can be seen by much lower execution time compared to the other trials. By entering the most efficient downstream controller in every control loop, FC$^3$ recognizes when subgoals are achieved and automatically moves to the next controller. 
\section{Conclusion}
We proposed a novel framework to generate a reactive hierarchical policy that exploits a direct interaction of discrete decision making with the underlying controllers.
The system coordinates control chains, thereby realizing reactive plan switching and controller selection, by reasoning over their feasibility in the current situation. 
FC$^3$ does not rely on sufficient symbols or require introducing additional symbols to coordinate these controllers, but utilizes geometric features and constraints to determine feasibility of controllers, chains and the task itself.


\bibliographystyle{IEEEtran}
\bibliography{IEEEabrv, root}



\end{document}